\documentclass[10pt,twocolumn,letterpaper]{article}
\usepackage{config/cvpr}     

\usepackage[dvipsnames]{xcolor}

\usepackage{times}
\usepackage{epsfig}
\usepackage{graphicx}
\usepackage{amsmath}
\usepackage{amssymb}
\usepackage{booktabs, nicematrix}
\usepackage{makecell}
\usepackage{multirow}
\usepackage{verbatim}
\usepackage{epsfig}
\usepackage{float}
\usepackage{wrapfig}
\usepackage[accsupp]{axessibility}  
\usepackage[hyphens]{url}
\usepackage{colortbl}
\usepackage{cuted}
\usepackage{bm}
\usepackage{xspace}
\usepackage{multirow}
\usepackage{tabularx}
\usepackage{xcolor}
\usepackage{soul}

\usepackage{stfloats}
\usepackage{tikz}

\usepackage{colortbl}
\usepackage{xcolor}
\usepackage{array}
\usepackage{diagbox}
\usepackage{pgfmath} 

\usepackage{pifont} 
\usepackage[normalem]{ulem}
\usepackage{arydshln}
\definecolor{cvprblue}{rgb}{0.21,0.49,0.74}

\usepackage[pagebackref,breaklinks,colorlinks,citecolor=cvprblue]{hyperref}

\usepackage{listings}
\definecolor{backcolour}{rgb}{0.95,0.95,0.92}
\lstdefinestyle{mystyle}{
    backgroundcolor=\color{backcolour},  
    keywordstyle=\scriptsize,
    basicstyle=\ttfamily\scriptsize,
    breakatwhitespace=false,
    captionpos=b,                    
    keepspaces=true,                 
    numbers=left,                    
    numbersep=5pt,  
    showspaces=false,                
    showstringspaces=false,
    showtabs=false,                  
    tabsize=4
}

\title{Moving by Looking: Towards Vision-Driven Avatar Motion Generation}

\newcommand{\methodname}{\textit{CLOPS}\xspace}

\newcommand{\mocap}{\mbox{MoCap}\xspace}

\newcommand{\threeD}{\xspace{3D}\xspace}

\newcommand{\qheading}[1]{\noindent\textbf{#1:}}

\newcommand{\MPrior}{D\xspace}  
\newcommand{\Q}{Q\xspace}

\definecolor{GreenColor}{rgb}{0.137,0.573,0.565}
\definecolor{OrangeColor}{rgb}{0.914,0.541,0.0.141}
\definecolor{PurpleColor}{rgb}{0.5,0,0.7}
\definecolor{BlueColor}{rgb}{0,0,1}

\newcommand{\supmatCOLOR}{black}

\newcommand{\supmat}{\textcolor{\supmatCOLOR}{{Sup.~Mat.}}\xspace}

\begin{document}
\author{
Markos Diomataris $^{1,2}$\quad
Mert Albaba$^{1,2}$\quad
Giorgio Becherini$^{1}$\quad
Partha Ghosh$^{1}$\quad
\\
Omid Taheri$^{1}$\quad
Michael J. Black$^{1}$\quad \\
\small
$^{1}$Max Planck Institute for Intelligent Systems, T{\"u}bingen, Germany \hspace{0.2in} 
$^{2}$ETH Z\"{u}rich, Switzerland\\
}

\twocolumn[{%
  \renewcommand\twocolumn[1][]{#1}%
 \maketitle
   \vspace*{-0.7cm}
   \begin{center}
    \centerline{ \includegraphics[trim=0mm 0mm 0mm 0mm, clip=true, width=1. \linewidth]{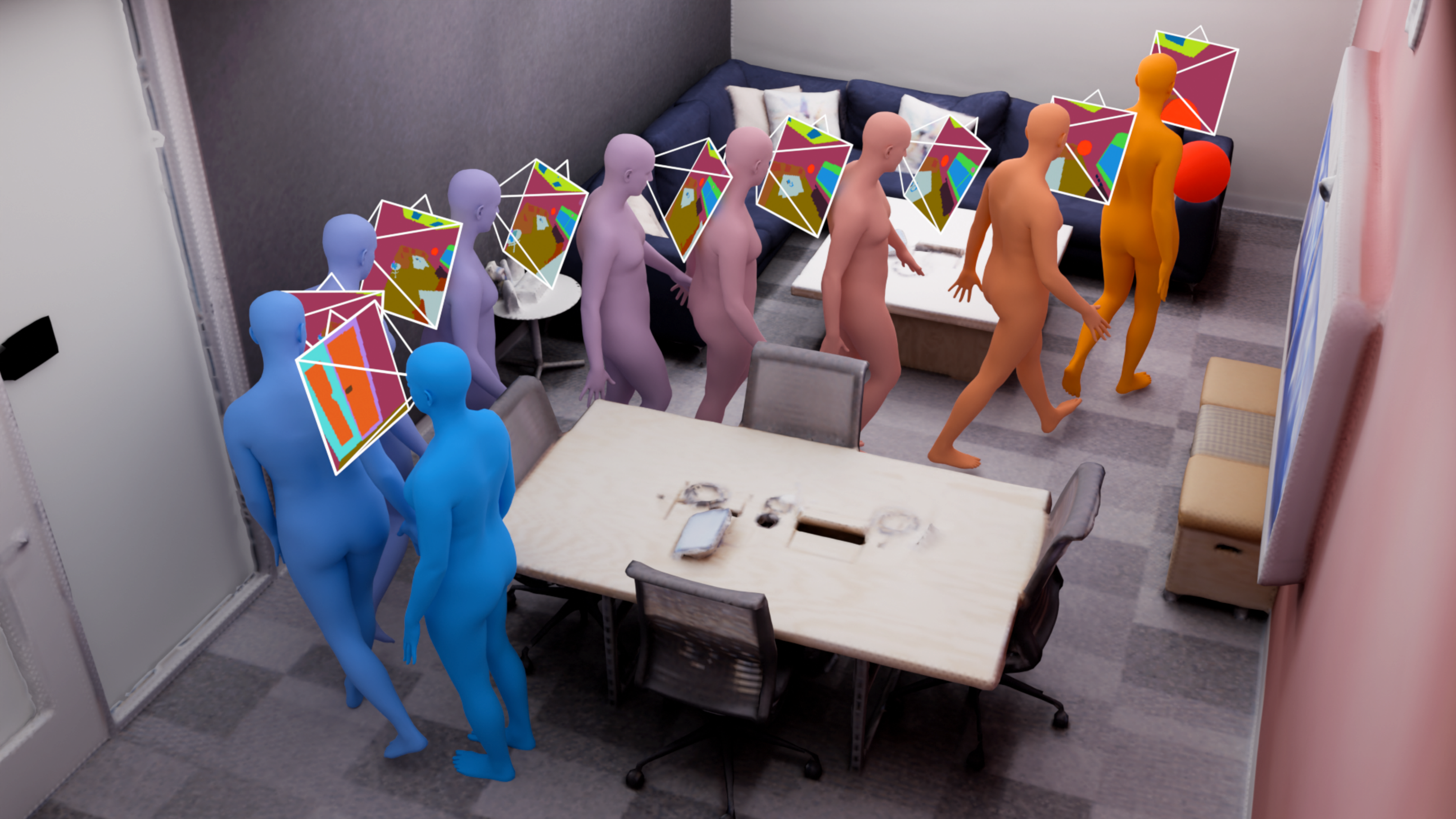}}
 \vspace*{-2.6em}
  \end{center}
  
  \begin{center}
\captionof{figure}{\textbf{\methodname} has learned to look for the goal (red sphere) using only its egocentric vision (here visualized as the vision cone in front of the avatar's head). It uses its ``eyes'' in order to discover where the goal is and navigate to it without colliding with the environment.}
\label{fig:teaser}
\end{center}%
}]

\begin{abstract}
The way we perceive the world fundamentally shapes how we move, whether it is how we navigate in a room or how we interact with other humans.
Current human motion generation methods, neglect this interdependency and use task-specific ``perception'' that differs radically from that of humans.
We argue that the generation of human-like avatar behavior requires human-like perception.
Consequently, in this work we present \methodname, the first human avatar that solely uses egocentric vision to perceive its surroundings and navigate. 
Using vision as the primary driver of motion however, gives rise to a significant challenge for training avatars: existing datasets have either isolated human motion, without the context of a scene, or lack scale.
We overcome this challenge by decoupling the learning of low-level motion skills from learning of high-level control that maps visual input to motion.
First, we train a motion prior model on a large motion capture dataset.
Then, a policy is trained using Q-learning to map egocentric visual inputs to high-level control commands for the motion prior. 
Our experiments empirically demonstrate that egocentric vision can give rise to human-like motion characteristics in our avatars. 
For example, the avatars walk such that they avoid obstacles present in their visual field.
These findings suggest that equipping avatars with human-like sensors, particularly egocentric vision, holds promise for training avatars that behave like humans. Code, models and data are available at our \href{https://markos-diomataris.github.io/projects/clops/}{project page}
\end{abstract}

\section{Introduction}
\label{sec:introduction}

Imagine looking for your keys. At every moment, your eyes provide visual information that is transformed to motor commands controlling your muscles that move your body.
This motion enables your eyes to move and gather more information that, in turn, drives your body.
All of this complex interplay of vision and motion results in the ``searching'' behavior that is driven by a very simple objective that is hard to formulate mathematically, i.e.~``find your keys.''

Human motion is inherently tied to human perception -- visual, tactile, proprioceptive, and auditory -- and here we focus on visual perception.
We prefer to walk facing forward so that our eyes can detect obstacles.
If we wear a blindfold, the way we move will drastically change: we will walk more slowly and stretch our hands to sense the world around us.
This interdependency of vision and body motion shapes ``human-like'' motion.
Fundamentally, ``we see in order to move; we move in order to see'' (William Gibson).

Even though vision plays such an important role in the way humans move, there is no human avatar motion generation system that is driven by human-like vision.
In the majority of cases, avatars perceive the world as abstract representations like waypoints \cite{wang2020adversarial}, trajectories \cite{SAMP2021}, point clouds \cite{PURPOSER2024}, or occupancy grids \cite{TRUMANS2024, LINGOMOTION2024}. 
Typically, the agent is omniscient, with full knowledge of the scene.
All of these types of sensors, while practical, are far from resembling how real humans perceive the world.
Inspired by how human vision shapes human motion, we propose \textbf{\methodname}: Controlling avatar Learning with an Observing Perceptual System. It is the first human avatar driven by human-like egocentric vision.
Our key goal is then to train an avatar to move using vision as a primary sensor.

To that end, we focus on the fundamental task of scene navigation, but with an important twist.
Instead of providing the avatar with the goal location it needs to reach, as commonly done in existing literature, the avatar must use its egocentric vision to identify the goal and navigate toward it.
We illustrate this task in Fig.~\ref{fig:teaser}.
More concretely, a human avatar has to navigate and reach a goal location (represented as a red sphere) while minimizing collisions with the scene.
Input to the motion generation system consists of egocentric images and the output is natural human motion.
\methodname has to learn how to translate visual input to motion, understand what obstacles look like, how to move to avoid them, and deal with the fact that it cannot observe the whole scene at once.
It not only needs to figure out how to go to the goal, but also discover the goal itself.

The first challenge that arises is finding the appropriate data to train our system.
We could either use an existing dataset with egocentric views captured in real scenes \cite{grauman2024egoexo4} or render the egocentric view of motions captured in virtual scenes \cite{TRUMANS2024, LINGOMOTION2024}.
This could give us pairs of human motion with egocentric vision.
Nevertheless, there are two important shortcomings with this approach.
First, it would result in a system that only works when deployed in scenes where the domain gap is small. 
Second, most \mocap data (e.g.~AMASS \cite{Mahmood2019-bi}) does not contain the scene and methods to automatically add a scene to \mocap data are limited \cite{yi2023mime}.

A more suitable learning paradigm, that circumvents the domain gap problem without constraining our dataset options is Reinforcement Learning (RL).
By using RL, we could directly place the avatar in any scene and, after gathering experience by moving in it, it would learn to navigate.
In this approach, a dataset along with a reward function would be used to define natural human motion, similar to \cite{Peng2021-vr, li2024egogen, Zhao2023Synthesizing}.
However, naively adopting the RL framework would give rise to another important challenge.
In this approach, where input is images and output is human poses, the policy would have to simultaneously learn how to map visual inputs to actions in order to navigate (i.e.~reach the goal without colliding) and which sequences of actions generate natural human motion. 
This added complexity makes training infeasible.

Both data-driven and end-to-end RL methods have their drawbacks for this task. 
Our key observation is that, by combining them, we can get the best out of both -- a data-efficient method that generalizes to new scenes.
First, we train a motion prior to efficiently learn the low-level details of how humans move (independent of any scene) from a large \mocap dataset.
Then, a policy uses Q-learning~\cite{qlearning2013} to only learn how to map the avatar's egocentric vision to a set of high-level controls that drive the pre-trained motion prior.
This decoupling reduces the state-action space of the RL problem, making it tractable.
In addition, it allows us to use any motion dataset to learn the low-level details of human motion.
This is loosely analogous to how a person would learn to play a video game.
The game's ability to make a game character walk forward is already built in (the motion prior) and the player’s task is to learn when to press the ``walk forward'' button based on what they observe on the screen (RL policy).

Specifically, \methodname consists of two neural networks: a conditional motion prior \MPrior and a state-action value function \Q.
\MPrior is a conditional Variational Auto Encoder (c-VAE) that autoregressively generates human motion. It is trained on motion data from AMASS \cite{Mahmood2019-bi} and its condition signal is the desired pose (translation and orientation) of the avatar's head. 
Its purpose is to generate natural human motion that aligns the avatar's head with the target pose given as a condition.
\Q is a state-action value function trained with Q-learning, where its action space is a discrete set of coordinate frames that conditions the motion generator \MPrior.

We train these models in two stages (as shown in Fig.~\ref{fig:model}).
First, the motion prior \MPrior is trained on AMASS. Then, we freeze its weights and train policy \Q using Q-learning. 
The \Q network learns to rank available head target poses, selecting the optimal one so that, when passed to \MPrior, it generates motion that navigates to the goal while avoiding collisions. 
To generate motion, egocentric information is first given to \Q.
This includes semantic segmentation, depth and the binary mask of the goal sphere.
Given this, \Q predicts the desired target pose for the avatar's head. Then, \MPrior autoregressively generates motion until it reaches the target head pose. This process is iterated until the avatar reaches the goal sphere. In Fig.~\ref{fig:teaser} we show all the frames of a motion where the policy \Q takes an action. We also visualize part of its input which is the semantic segmentation of the egocentric view inside the frustums.

In summary, we present \methodname, a novel approach that enables human avatars to navigate in scenes using egocentric vision.
We demonstrate that \methodname shows preference to look in the direction it walks, looks around to find the goal and keeps it in its field of vision while moving towards it.
These behaviors emerge ``for free'' as a result of egocentric vision being the driver of motion.
To the best of our knowledge, this is the first method that equips human avatars with a virtual ``eye'' and explores the benefits of human-like perception.
With this work, we hope to inspire future research to consider how human-like sensors act as a perceptual bottleneck from which natural human behavior emerges.
Data and models will be available for research purposes.

\section{Related Work}
\label{sec:related}

A key research goal is to develop learning processes that capture the nuances of human movement and behavior and then generate motions that are  indistinguishable from the real ones.
Human motion generation, however, is a complex sequence-prediction problem and hence datasets~\cite{Mahmood2019-bi, PROX:2019,BABEL:CVPR:2021,GRAB:2020, wang2022humanise, SAMP2021, jiang2024autonomous} play a pivotal role in enabling realism and diversity. These datasets contain motions captured from real humans, usually interacting with scenes. They are often accompanied by text that describes the action being performed by the actor. They enable learning a mapping from high level inputs such as text descriptions \cite{Tevet2022-ab}, keyframes \cite{DART2024} or waypoints \cite{diomataris2024wandr, TRUMANS2024, Zhang_2022_CVPR, Zhao2023Synthesizing} to full-body human motion. The underlying common objective often revolves around generating a high level of user controllability or a high level of automation. 
Here, we focus on three broad categories of research on human motion generation.

\subsection{Sensors of Avatars}


The generation of human motion in scenes requires that the agent is aware of its environment and can avoid obstacles and interact with objects.
A straightforward approach to obstacle avoidance uses precomputed waypoints and places such that, when the avatar moves from one way-point to the next, it does not collide with anything \cite{SAMP2021,zhang2024scenic}. 
%
To avoid pre-computing waypoints, methods use virtual sensors that make the motion generator aware of its surroundings. For example, \cite{TRUMANS2024, jiang2024autonomous,Starke2019-bl} use a 3D occupancy grid that surrounds the avatar and informs it about the surrounding objects in very close proximity. To navigate large distances, such methods typically still need to define way-points using A* algorithm. 
Several methods go beyond avoiding obstacles to interacting with them, e.g.~sitting on objects \cite{Zhao2023Synthesizing,Starke2019-bl,SAMP2021}.
In \cite{PURPOSER2024}, the authors use a pointcloud representation of the scene instead of 3D voxels. 
Li et al.~\cite{li2024egogen} use Lidar-like one-dimentional depth information as the sensor input. They define an explicit reward that guides the agent to turn its head towards the goal as well as walk towards it. This emphasizes the need for reward shaping to make the emerging behaviors look human.
Such reward shaping, when done manually, is extremely difficult. 
As a consequence, the generated motion looks ``lifeless'' to the human eye. 

The existing sensors above either access oracle information about the scene (e.g.~point-clouds) or sense very sparse locations of where to walk next. These sensors, unfortunately by design lack a realistic connection to the avatar's body. 
Some work recognizes this problem and engineers ways, either with rewards \cite{li2024egogen} or condition signals \cite{diomataris2024wandr}, to force the avatar to align its gaze with the goal, like real humans do.
In contrast to prior work, we give the agent only the input from a monocular ``vision like'' sensor. 
Our hypothesis is that having a sensor that is closer to human vision will result in motions that are more human-like.

\subsection{Reinforcement Learning for Motion}
There are two major categories of RL based methods that learn to move like humans do:

\qheading{Low level controller}
RL learns the low-level control in simulation \cite{tessler2024maskedmimic, yao2024moconvq, tevet2024closd}. In these cases, RL agents are trained with dense reward functions (per frame keypoints, waypoints) and learn the distribution of human motion, either with adversarial \cite{Peng2021-vr} or imitation \cite{tessler2024maskedmimic} rewards. In these cases, the purpose of RL is to map the ``in vitro'' avatar motion to a physically simulated humanoid body.

\qheading{High-level controller}
Another approach uses RL to learn to control a pretrained motion generator (i.e.~motion prior) using its latent space (\cite{Zhang_2022_CVPR, li2024egogen, Zhao2023Synthesizing, DART2024}). Here, RL effectively explores motion primitives with the goal of maximizing a reward function. A downside of this approach is that the RL algorithm has a large action space. Namely, the action space  is as big as the latent space of the motion prior. This complicates the construction of the reward function, often requiring specific rewards to penalize things like foot sliding or unnatural poses.
In contrast to previous approaches, we drastically reduce the size of the action space of the RL controller by discretizing the action space.
We only specify the target head poses rather than target joint locations and poses of the entire body. 

\subsection{Controlling Human Motion}
 The most popular approach to control human motion generation is through text input
 \cite{Petrovich2022-ie, Athanasiou2023-nd,athanasiou2024motionfix}.
Such methods provide little to no ``autonomy'' to the agent.
In contrast, another line of work \cite{li2024egogen, diomataris2024wandr} creates agents that require minimal input from the user and may support perpetual motion.
These works usually lack the level of semantic control of the text-to-motion approaches but result in avatars with more agent-like properties, \ie deciding what actions to take on their own. 
Our work extends the scope of the second body of work. \methodname operates fully autonomously and requires no human intervention. By solely using egocentric vision it learns to navigate in a scene without the need of waypoints or keyposes.

\section{Method}
\label{sec:method}

\begin{figure*}[t]

    \centering
    \includegraphics[width=0.9\textwidth, trim=0cm 0.0cm 0cm 0cm, clip=true]{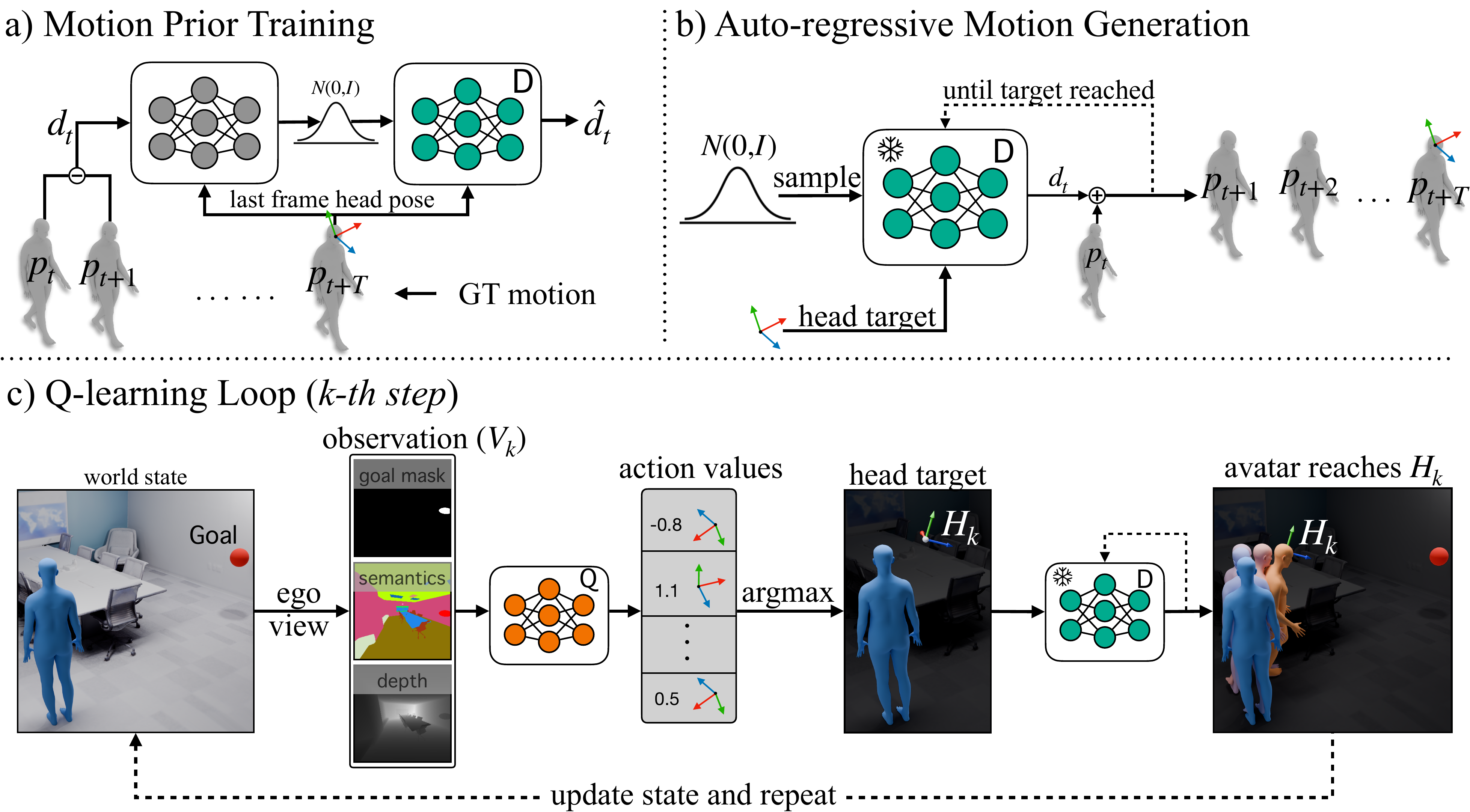}
    
    \caption{\textbf{\methodname architecture}. (a) The motion prior VAE to learns the distribution of pose deltas conditioned on the head pose of a future sequence frame. (b) The decoder \MPrior can be used to autoregressively generate motion able to reach the conditioning head pose target. (c) To create a motion chunk, the policy \Q uses the avatar's ego-view to predict a head target pose that is then passed to the motion prior to generate a motion chunk. This is iterated until the goal is reached.}
    \label{fig:model}

\end{figure*}

Our goal is to create a human avatar that can autonomously navigate to a goal in a scene while relying only on its egocentric vision. We choose the goal to be a red sphere, as shown in Fig.~\ref{fig:teaser}, and its location is \textit{not} explicitly given to the avatar. An implicit task of our agent is also to discover the goal visually.
We design \methodname as an iterative process. At every iteration $k$, an egocentric visual observation $V_k$ is provided as input.
For each observation $V_k$, a sequence $P_k=[p^k_t, ..., p^k_{t+T}]$ of $T$ SMPL-X \cite{smplifyPP} body poses is autoregressively generated. We call $P_k$ a motion chunk. We use $k$ to refer to the observation loop and $t$ to the nested motion generation loop (see Fig.~\ref{fig:model}c). 

We define $V_k \in \mathbb{R}^{64 \times 64 \times 5}$ as the channel-wise concatenation of the egocentric depth map (1-channel), RGB semantic segmentation (3-channels) and binary mask (1-channel) of the goal sphere (visualized in Fig.~\ref{fig:model}c). The avatar's field of view is $130^o$.
Our objective can now be formulated as learning a mapping 
\begin{equation}
f: V_k \rightarrow P_k
\label{eq:task}
\end{equation}
so that the generated output $\mathbf{P} = [P_1, P_2, ...]$, comprised of all the generated motion chunks, is a natural human motion that navigates to reach the goal without colliding with obstacles.

We decompose learning the mapping of Eq.~\ref{eq:task} into two sub-problems.
First, our method trains a motion prior that, given a target head pose $H_k$, generates full-body motion to reach it in a natural way:
\begin{equation}
P_k = \MPrior(H_k)
\label{eq:movetosee}
\end{equation}
where $H_k \in \mathbb{R}^{4 \times 4}$ describes the target translation and orientation of the head in matrix format.
Note that the agent does not have independent eye control and can only explore its world by translating and rotating its head to acquire new image information.
Thus, we focus on head motion because it is what allows exploration; the body simply follows the head.
We call this sub-problem \textit{moving to see} and we learn it from data (Sec.~\ref{sec:move_to_see}).
Second, our method learns how to map egocentric observations to the proper target head poses $H_k$ that will guide the avatar to the goal:
\begin{equation}
H_k = Q(V_k)
\label{eq:seetomove}
\end{equation}
We call this sub-problem \textit{seeing to move} and we learn it using Q-learning~\cite{qlearning2013} (Sec.~\ref{sec:see_to_move}).
The composition of Equations~\ref{eq:movetosee} and~\ref{eq:seetomove} constitutes a system that, given egocentric vision input, produces a natural human motion chunk:
\begin{equation}
P_k = \MPrior(\Q(V_k)).
\label{eq:taskcomposition}
\end{equation}

In Fig.~\ref{fig:model}c we show how Eq.~\ref{eq:taskcomposition} is implemented.
First, the avatar's egocentric observations $V_k$ are fed to the \Q network.
This in turn, predicts a score (value) for a discrete set of possible next target head poses. The one that ranks the highest, $H_k$, is chosen and given to the motion prior.
Finally, the motion prior \MPrior autoregressively generates motion so that the avatar naturally moves and aligns its head with the target $H_k$ chosen by the \Q network.
This process is iterated until the avatar reaches the goal sphere. We provide a visual break down in the \textbf{supplementary video}.

\subsection{Learning to Move from Data}
\label{sec:move_to_see}
The role of the motion prior is to generate a motion chunk $P_k$ that reaches  the specified target head poses $H_k$.
We train it on 3D human motions from AMASS \cite{Mahmood2019-bi}.
The design is based on the autoregressive motion prior proposed in WANDR \cite{diomataris2024wandr}.
At its core, a VAE is trained to learn the distribution of the differences between two consecutive poses $p_t$ and $p_{t+1}$ (see Fig.~\ref{fig:model}a).
We call these differences pose deltas $d_t$.
Both the encoder and decoder are conditioned on the head pose of a future body at timestep $t+T$.
This effectively makes the learned distribution capture pose deltas that, when integrated, produce motion that aligns the avatar's head with the conditioned head target pose.
Then, by sampling the latent space and integrating the predicted deltas, motion is generated (Fig.~\ref{fig:model}b).
Intuitively, $T$ controls how controllable the prior is in terms of reaching the head target $H_k$. Bigger $T$ results to smoother motion but the avatar's head might not reach $H_k$ after $T$ frames. Smaller $T$ on the other hand makes the prior more responsive but the overall motion becomes more abrupt. We found $T=1sec$ to strike a good balance.  

A problem that arises, however, is that we do not know exactly how many frames need to be generated in order for the avatar's head to reach $H_k$. For example, to move one meter forward, more time is needed when starting still compared to when the avatar is already walking. This is the reason why we design the motion prior as an autoregressive process.
The method can generate poses until the avatar's head reaches the target pose, within some threshold, or a maximum number of poses $T$ has been generated.
Now that we can control the avatar to move its head and effectively observe the scene from a desired viewpoint $H_k$, we next concentrate on designing a mechanism that can generate a sequence of $H_k$'s to complete the task -- find and reach the goal.

\subsection{Learning to See from Experience}
\label{sec:see_to_move}
To learn the mapping \Q  from egocentric visual input $V_k$ to target goal poses $H_k$, as expressed in Eq.~\ref{eq:seetomove}, we use Q-learning.
A challenge that arises is that, since $H_k$ is a continuous valued variable, applying Q-learning is not straight forward. To overcome this, we observe that, for the avatar to navigate in a scene without colliding, a sufficiently large set of pre-defined target head poses is enough. We use AMASS to compute the difference between any two head poses that are $T$ frames apart from each other (here $T=1\text{sec})$. Then, we cluster these head pose deltas to get $N$ distinct head pose deltas that represent the $N$ actions available to the \Q network. For example, some of these actions will move the avatar forward, others will make it turn or step sideways.

To train the \Q network we use the following reward function:
\begin{equation}
    r =
    \begin{cases}
        +r_r, & \text{if goal reached AND visible} \\
        -r_t - r_c - r_m, & \text{otherwise} 
    \end{cases}
\end{equation}
where $r_t$ is a constant time penalty, $r_c$ penalizes collision, and $r_m$ penalizes the agent if it stays at the same place without moving.
The time penalty pushes the agent to efficiently go to the goal, since the sooner the movement terminates, the fewer $r_t$'s are accumulated.
The collision penalty incentivizes avoiding scene penetration.
In cluttered scenes however, the agent might learn in early training that it is better to not move at all to avoid collisions. If this happens, then the probability of discovering and then  going to the goal drops dramatically.
The $r_m$ term mitigates this by incentivizing the agent to explore the space around it. It increases the chances of, at first randomly, stumbling on the goal and later more successfully learning to visually search, recognize and move towards it.

We reward the agent with $r_r$ when it manages to reach a state where the goal is closer than $50$ cm \textit{and} is within its field of vision.
Notice that this term captures the essence of what finding an object means. This is one of the benefits of using vision as input since it allows the definition of rewards in the ego-view space.

Note that we do not provide any explicit reward to the avatar defining if the motion was good, which path to follow, where to look, what the goal looks like, or if an action moved it closer to or further away from the goal. The \Q network learns all of this through exploration using only egocentric observations.


\begin{figure*}[!h]
\centerline{\includegraphics[width=\textwidth]{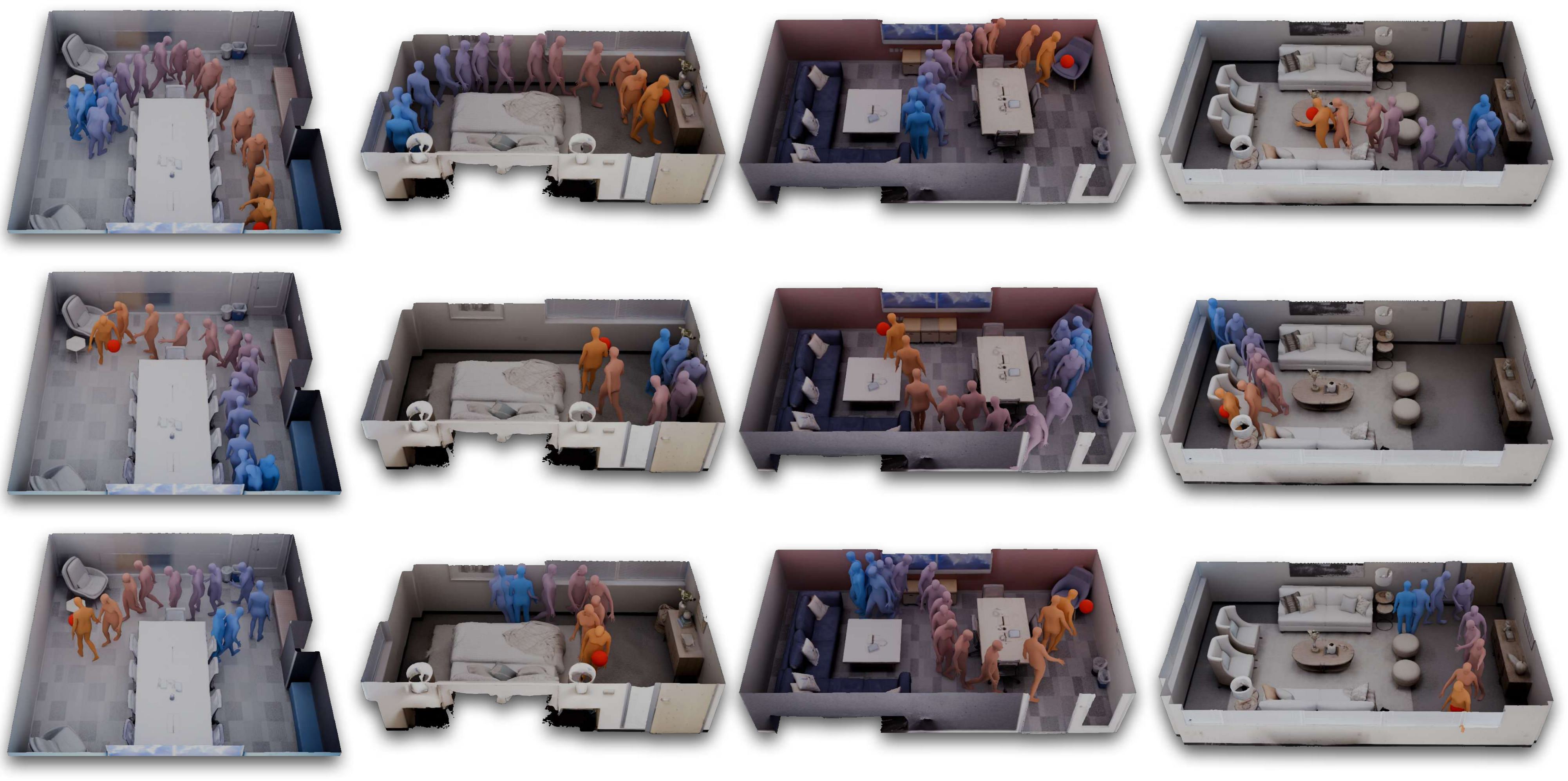}}
    \caption{Example episodes in different scenes. The red ball is the target. The avatar's starting pose is in blue and the ending pose in orange. Note how the avatar orients itself to identify the goal and moves to avoid obstacles, purely using visual input. See the \textbf{supplementary video} for visualization of the motion quality.}
    \label{fig:quals}
    
\end{figure*}
\section{Experiments}
\label{sec:experiments}
Our experiments aim to answer three main questions: (1) How effectively can \methodname learn to visually discover the goal and reach it without colliding? (2) Does it generalize when tested on scenes it has never seen during training? (3) Does the position of the ego-view sensor on the head affect how \methodname learns to move?

First, our method trains the motion prior $\MPrior$ on the AMASS dataset. We filter out sequences containing actions unrelated to navigation, such as dancing, kickboxing, \etc.
It takes approximately 15 hours to train the prior on a single NVIDIA-A100 GPU.
For all the different reported experiments, this motion prior remains the same and only the policy $\Q$ is retrained.
We follow common practices used in Q-learning such as double q-learning \cite{hado2016double} and prioritized experience replay \cite{schaul2015prioritized}.
We train the policy until the average predicted value converges, which is about $30 K$ steps and takes around $10$ hours on a single NVIDIA-A100 GPU. We include all the details and hyperparameters in 
\supmat

To train and evaluate \methodname, we use scenes from the Replica dataset~\cite{straub2019replica}.
We pick five scenes, here referred to as $S1$ to $S5$, that are diverse in terms of layout, furniture, and complexity. Some of them have wider pathways and require mostly long-term planing, whereas others are more cluttered with obstacles and require more fine-grained control. See \supmat for more details.
We refer to a complete motion sequence as an ``episode.''
For every episode, we randomize the avatar's starting position and orientation as well as the goal's $x,y$ location.
Termination occurs either after 15 seconds of generated motion or if the avatar reaches the goal earlier.
Since $T$ is $1$ second, each episode is comprised of at least $15$ actions. The episodes do not terminate when a collision happens. We found that terminating on collision dramatically limits the agent's gathered experience and results in poor performance. Throughout training, the reward function remains the same.

To evaluate \methodname we use metrics that reflect the method's ability to navigate to the goal without colliding while also producing good quality motion:
\begin{itemize}
    \item \textbf{Success Rate (SR)} measures the average percentage of episodes in which \methodname successfully reaches a state where the goal is within $50$ cm of the human and within its field of view. 
    \item \textbf{Collision Rate (CR)} measures the average percentage of actions that result in collision with an obstacle. It is a strict metric since small penetrations with the scene are weighted as much as bigger ones.
    \item \textbf{Foot Skating (FS)} reflects the quality of the motion. We adopt the definition in \cite{araujo2023circle, diomataris2024wandr}. It is the average percentage of frames where the lowest vertex of the human mesh moves more than 0.66cm relative to the ground.
\end{itemize}
Every reported metric is a result of averaging $500$ episodes where the initial position and orientation of the human as well as the goal are randomized.

\subsection{Generalization of \textbf{\methodname}}
\label{sec:generalization}




\newcommand{\maxnumA}{100}  
\newcommand{\minnumA}{65}    

\newcommand{\maxnumB}{50}   
\newcommand{\minnumB}{9}   

\newcommand{\applycolorA}[1]{%
    \pgfmathparse{100.0 * (#1 - \minnumA) / (\maxnumA - \minnumA)}%
    \xdef\colorval{\pgfmathresult}%
    \cellcolor{orange!\colorval} #1
}

\newcommand{\applycolorB}[1]{%
    \pgfmathparse{100.0 * (#1 - \maxnumB) / (\minnumB - \maxnumB)}%
    \xdef\colorval{\pgfmathresult}%
    \cellcolor{orange!\colorval} #1
}

\newcolumntype{W}{!{\color{white}\vrule width 2pt}}
\begin{table*}[h]

\centering
\renewcommand{\arraystretch}{1.4}
\setlength{\tabcolsep}{10pt}

\begin{subtable}{0.5\linewidth}

\centering
\textbf{\quad \quad \quad \quad $\uparrow$ Success Rate (\%)} \\[5pt]
\resizebox{0.95\columnwidth}{!}{

\begin{tabular}{|c | c W c W c W c W c|}
\scalebox{0.75}{
    \tikz{
        \node[below left, inner sep=1pt] (def) {Train};%
        \node[above right,inner sep=1pt] (abc) {Test};%
        \draw (def.north west|-abc.north west) -- (def.south east-|abc.south east);
    }
}
&
S1 &
S2 &
S3 &
S4 &
S5 \\

\hline
        
S1 & \applycolorA{93} & \applycolorA{84} & \applycolorA{96} & \applycolorA{78} & \applycolorA{85} \\
\addlinespace[2pt]
S2 & \applycolorA{78} & \applycolorA{87} & \applycolorA{93} & \applycolorA{73} & \applycolorA{78} \\
\addlinespace[2pt]
S3 & \applycolorA{81} & \applycolorA{72} & \applycolorA{84} & \applycolorA{72} & \applycolorA{72} \\
\addlinespace[2pt]
S4 & \applycolorA{90} & \applycolorA{84} & \applycolorA{94} & \applycolorA{78} & \applycolorA{73} \\
\addlinespace[2pt]
S5 & \applycolorA{87} & \applycolorA{78} & \applycolorA{90} & \applycolorA{66} & \applycolorA{68} \\
            
\end{tabular}}

\end{subtable}%
\hfill
\begin{subtable}{0.5\linewidth}

\centering
\textbf{\quad \quad \quad \quad $\downarrow$ Collision Rate (\%)} \\[5pt]
\resizebox{0.95\columnwidth}{!}{

\begin{tabular}{|c | c W c W c W c W c|}

\scalebox{0.75}{
    \tikz{
        \node[below left, inner sep=1pt] (def) {Train};%
        \node[above right,inner sep=1pt] (abc) {Test};%
        \draw (def.north west|-abc.north west) -- (def.south east-|abc.south east);
    }
} &
S1 &
S2 &
S3 &
S4 &
S5 \\

\hline

S1 & \applycolorB{26} & \applycolorB{16} & \applycolorB{35} & \applycolorB{42} & \applycolorB{40} \\
\addlinespace[2pt]
S2 & \applycolorB{42} & \applycolorB{10} & \applycolorB{35} & \applycolorB{43} & \applycolorB{43} \\
\addlinespace[2pt]
S3 & \applycolorB{40} & \applycolorB{20} & \applycolorB{19} & \applycolorB{45} & \applycolorB{38} \\
\addlinespace[2pt]
S4 & \applycolorB{39} & \applycolorB{20} & \applycolorB{35} & \applycolorB{38} & \applycolorB{38} \\
\addlinespace[2pt]
S5 & \applycolorB{39} & \applycolorB{17} & \applycolorB{36} & \applycolorB{40} & \applycolorB{27} \\
            
\end{tabular}}

\end{subtable}
    \caption{We trained \methodname five times, each on a different scene (S1 $\rightarrow$ S5), and tested each version separately on every scene. We present the results as confusion matrices. Our method is able to find and navigate to goals (Success Rate) while doing a good job of avoiding obstacles (Collision Rate). The non-diagonal structure of the matrices suggest that \methodname generalizes these skills to new scenes.}
    \label{tab:double_confusion}

\end{table*}

There are two different kinds of generalization to be evaluated.
The first one tests \methodname's ability to generalize to new initial states in a scene it was trained on. The second tests how well it can perform in a scene it has never seen before.
We train five policies, one on each scene, and test each on all five scenes.
We report the results for Success Rate and Collision Rate in Table~\ref{tab:double_confusion} as two confusion matrices. 
Rows correspond to the training scene while columns to the test scene. 
Color intensity is proportional to better performance (higher Success Rate or lower Collision Rate).

In general, we notice that the avatar manages to keep a high Success Rate in all scenes, both seen and unseen, with scores no less that $66\%$. 
This means that the policy is able to generalize the skill of looking for and going to the goal even in unseen scenes.

One might expect the confusion matrices to be strongly diagonal. Interestingly, this is not the case. We observe that, for the most part, performance is dependent on the test scene (column) rather than on the training scene (row). This suggests that, whatever the policy learns to visually recognize and avoid obstacles, is transferred to new scene layouts. It is the test scene's complexity that primarily determines the performance, rather than the training.
Indeed, scenes with cluttered obstacles, like $S4, S5$, make it hard for the avatar to navigate, hence the high collision rates in those columns. 

\subsection{Comparison with EgoGen}
\label{sec:comparison}
\begin{table}[tb]
\centering
\begin{tabular}{l|ccc}
\Xhline{2\arrayrulewidth}
\textbf{Method} & \textbf{SR $\uparrow$} & \textbf{Col $\downarrow$} & \textbf{FS $\downarrow$} \\
\hline
EgoGen~\cite{li2024egogen} (known Goal) & 48\% & 31\% & 40\% \\
\methodname (only Vision) & \textbf{68}\% & \textbf{27}\% & \textbf{27}\% \\
\hline
\methodname + (known Goal) & 100\% & 36\% & 26\% \\
\Xhline{2\arrayrulewidth}
\end{tabular}
\caption{Comparison with EgoGen on target reaching Success Rate (SR), Collision Rate (Col), and Foot Skating (FS). Despite relying solely on vision, our method outperforms EgoGen. Explicit goal information ensures \methodname always finds it.}
\vspace{-0mm}
\label{tab:egogen}
\end{table}
The method closest to \methodname is EgoGen~\cite{li2024egogen}. To our knowledge, it is the only method in the literature that generates avatar motion without providing some form of a path to follow.
Their method consists of a policy trained with proximal policy optimization~\cite{schulman2017proximal} to control the continuous space of a motion prior.
Their avatar uses a 1D lidar-like sensor that scans in front of the avatar. EgoGen also accepts as input the exact location of the goal in \threeD space.
They impose a set of rewards in order for the motion to look realistic, reduce foot skating, force the avatar to orient towards the goal, go near it, turn its head towards it, and avoid obstacles.
In contrast, \methodname is not given the location of the goal and only relies on vision to find it. Furthermore, our rewarding scheme is much simpler since there is no term related to motion realism like foot skating or penalty for out of distribution poses.
The Q network in \methodname has very little interference with the motion quality when compared to EgoGen where the policy controls the continuous space of a motion prior.
We train and test our policy on $S5$, matching exactly EgoGen's overall training and testing framework, and present the results in Table~\ref{tab:egogen}.

\methodname is able to perform better in terms of reaching the goal while colliding less, even though it relies only on vision. This highlights the efficiency of our method, as it is trained with a sparser reward signal and a significantly higher-dimensional state space.
We qualitatively observe that EgoGen can sometimes better ``squeeze'' through narrow spaces. This is because its policy controls the full body of the avatar. Nevertheless, as soon as the avatar is not initialized in a way where the goal is close and in front of it, the method fails.

This comparison naturally brings the following question: what if \methodname knew where the goal is? In order to answer this, we train while providing the goal's location as an additional input to the egocentric vision. As we can see in the third row of Table~\ref{tab:egogen}, the success rate saturates to $100\%$, while collisions increase, presumably due to the method relying less on visual information.

\subsection{Qualitative Results}
\label{sec:qual}
Figure \ref{fig:quals} illustrates how \methodname navigates different scenes.
When starting from a state where the goal is not visible, it will try to move around and turn in order to detect it. Once the goal is seen, it tries to keep it in its field of view while moving towards it. We provide more qualitative results in the \textbf{supplementary video}. 

\subsection{Does Sensor Placement Influence Motion?}
\label{sec:sensor_ablation}
\begin{figure}

    \centering
    \includegraphics[width=0.45\textwidth, trim=0cm 0cm 0cm 0cm]{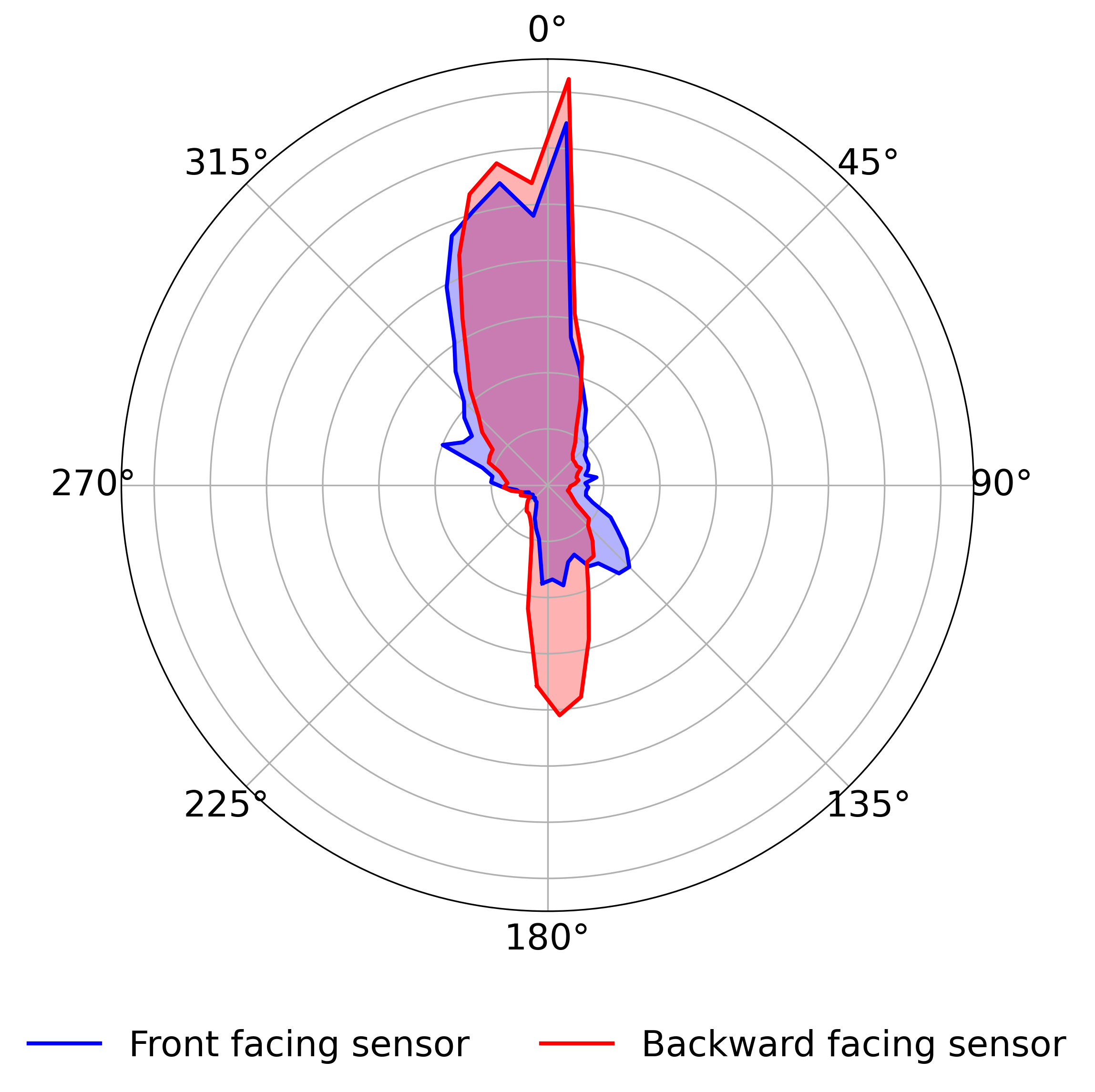}
    
    \caption{Placing the avatar's visual sensor facing backward forces \methodname's policy to change behavior and walk more backwards (since this is now its observation direction) in order to avoid obstacles.}
    \label{fig:angle_distribution}
    
\end{figure}

One question that still remains is whether the positioning of the ego-view sensor influences avatar motion? To investigate this, we design a counterfactual experiment. Let $\vec{f}$ be the natural front facing direction of the avatar's head (\ie where the nose points at) and $\vec{v}$ the avatar's pelvis velocity. Measuring the angle $\angle(\vec{f}, \vec{v})$ across $500$ motions generated by \methodname (blue distribution in Fig.~\ref{fig:angle_distribution}) reveals that, since the distribution is centered around zero, these two vectors mostly align \ie the avatar walks towards its observing direction.
If we now reverse the visual sensor (facing towards $-\vec{f}$) and retrain \methodname we observe a shift in the angle distribution of $\angle(\vec{f}, \vec{v})$ (red).
This shift tries to align the avatar's walking direction with its observing direction.
We attribute this to the fact that the only way for the avatar to systematically avoid obstacles and move towards the goal is to correlate its actions with the reward it receives. 
If the avatar does not look where it is walking, then the collision penalty becomes uncorrelated with the visual input and, thus, no useful learning takes place. In both cases, however, a large portion of the actions is forward-facing. 
The reason behind this is that the motion prior is trained on a dataset where backward-walking is very uncommon. 
Despite this, the change in the distribution in Fig.~\ref{fig:angle_distribution} suggests that the location of the sensor does impact the avatar motion.


\section{Limitations and Future Work}
Because \methodname does not have memory, it is not able to systematically search a room. If, while looking at the goal, it makes a motion that puts the goal out of its field of view, then it needs to find it all over again.
This can sometimes create scenarios where the avatar spends the whole episode looking for the goal in a corner of a room. Even though a fully-fledged memory system is out of the scope of this paper, we tried mitigating this with frame-stacking, a common strategy in Q-learning, but we did not observe any improvements. We hypothesize that works like 3D-MEM \cite{yang20253dmem3dscenememory} or, more generally, vision-language models that can act as a memory, are promising directions for the future research.

Another source of failure is the fact that the Q-learning policy can only control the avatar's head. This makes navigation in cluttered scenes challenging, as we observed in our experiments. Since there is no explicit control of the legs or the rest of the body, maneuvering through narrow space is difficult. This is made even more challenging by the fact that the avatar operates on 2D input instead of explicit 3D representations. There seems to exist a trade-off between the efficiency of policy learning and motion controllability. The more a policy can control, the more it needs to learn. 

\section{Conclusion}
In this work, we introduce \methodname, the first avatar that moves driven by egocentric vision. 
We extend the task of scene navigation beyond reaching a goal to include the process of discovering it through vision.
To address the lack of available data, we decouple the learning of motion skills from that of visual sensing. Low-level motion skills are learned in advance from data, while reinforcement learning is used to map visual inputs to motion-controlling commands.
Our experiments demonstrate that \methodname effectively learns to navigate and generalizes its skills to unseen environments. Furthermore, we show that the way human-like avatars perceive their surroundings directly influences how they move. We argue that human-like sensors are a crucial missing component in current avatar motion generation methods. 
With \methodname, we demonstrate that integrating vision into avatar motion generation is beneficial both for advancing human avatar capabilities and motion realism.
%


{\small
\bibliographystyle{config/ieeenat_fullname}
\bibliography{paper/ms}

\begin{thebibliography}{36}
\providecommand{\natexlab}[1]{#1}
\providecommand{\url}[1]{\texttt{#1}}
\expandafter\ifx\csname urlstyle\endcsname\relax
  \providecommand{\doi}[1]{doi: #1}\else
  \providecommand{\doi}{doi: \begingroup \urlstyle{rm}\Url}\fi

\bibitem[Ara{\'u}jo et~al.(2023)Ara{\'u}jo, Li, Vetrivel, Agarwal, Wu, Gopinath, Clegg, and Liu]{araujo2023circle}
Joao~Pedro Ara{\'u}jo, Jiaman Li, Karthik Vetrivel, Rishi Agarwal, Jiajun Wu, Deepak Gopinath, Alexander~William Clegg, and Karen Liu.
\newblock Circle: Capture in rich contextual environments.
\newblock In \emph{Proceedings of the IEEE/CVF Conference on Computer Vision and Pattern Recognition}, pages 21211--21221, 2023.

\bibitem[Athanasiou et~al.(2023)Athanasiou, Petrovich, Black, and Varol]{Athanasiou2023-nd}
Nikos Athanasiou, Mathis Petrovich, Michael~J Black, and Gül Varol.
\newblock Sinc: Spatial composition of 3d human motions for simultaneous action generation.
\newblock \emph{{arXiv}:2304.10417}, 2023.

\bibitem[Athanasiou et~al.(2024)Athanasiou, Ceske, Diomataris, Black, and Varol]{athanasiou2024motionfix}
Nikos Athanasiou, Alp{\'a}r Ceske, Markos Diomataris, Michael~J. Black, and G{\"u}l Varol.
\newblock {MotionFix}: Text-driven 3d human motion editing.
\newblock In \emph{SIGGRAPH Asia 2024 Conference Papers}, 2024.

\bibitem[Diomataris et~al.(2024)Diomataris, Athanasiou, Taheri, Wang, Hilliges, and Black]{diomataris2024wandr}
Markos Diomataris, Nikos Athanasiou, Omid Taheri, Xi Wang, Otmar Hilliges, and Michael~J. Black.
\newblock {WANDR}: Intention-guided human motion generation.
\newblock In \emph{Proceedings IEEE Conference on Computer Vision and Pattern Recognition (CVPR)}, 2024.

\bibitem[Grauman et~al.(2024)Grauman, Westbury, Torresani, Kitani, Malik, Afouras, Ashutosh, Baiyya, Bansal, Boote, Byrne, Chavis, Chen, Cheng, Chu, Crane, Dasgupta, Dong, Escobar, Forigua, Gebreselasie, Haresh, Huang, Islam, Jain, Khirodkar, Kukreja, Liang, Liu, Majumder, Mao, Martin, Mavroudi, Nagarajan, Ragusa, Ramakrishnan, Seminara, Somayazulu, Song, Su, Xue, Zhang, Zhang, Castillo, Chen, Fu, Furuta, Gonzalez, Gupta, Hu, Huang, Huang, Khoo, Kumar, Kuo, Lakhavani, Liu, Luo, Luo, Meredith, Miller, Oguntola, Pan, Peng, Pramanick, Ramazanova, Ryan, Shan, Somasundaram, Song, Southerland, Tateno, Wang, Wang, Yagi, Yan, Yang, Yu, Zha, Zhao, Zhao, Zhu, Zhuo, Arbelaez, Bertasius, Crandall, Damen, Engel, Farinella, Furnari, Ghanem, Hoffman, Jawahar, Newcombe, Park, Rehg, Sato, Savva, Shi, Shou, and Wray]{grauman2024egoexo4}
Kristen Grauman, Andrew Westbury, Lorenzo Torresani, Kris Kitani, Jitendra Malik, Triantafyllos Afouras, Kumar Ashutosh, Vijay Baiyya, Siddhant Bansal, Bikram Boote, Eugene Byrne, Zach Chavis, Joya Chen, Feng Cheng, Fu-Jen Chu, Sean Crane, Avijit Dasgupta, Jing Dong, Maria Escobar, Cristhian Forigua, Abrham Gebreselasie, Sanjay Haresh, Jing Huang, Md~Mohaiminul Islam, Suyog Jain, Rawal Khirodkar, Devansh Kukreja, Kevin~J Liang, Jia-Wei Liu, Sagnik Majumder, Yongsen Mao, Miguel Martin, Effrosyni Mavroudi, Tushar Nagarajan, Francesco Ragusa, Santhosh~Kumar Ramakrishnan, Luigi Seminara, Arjun Somayazulu, Yale Song, Shan Su, Zihui Xue, Edward Zhang, Jinxu Zhang, Angela Castillo, Changan Chen, Xinzhu Fu, Ryosuke Furuta, Cristina Gonzalez, Prince Gupta, Jiabo Hu, Yifei Huang, Yiming Huang, Weslie Khoo, Anush Kumar, Robert Kuo, Sach Lakhavani, Miao Liu, Mi Luo, Zhengyi Luo, Brighid Meredith, Austin Miller, Oluwatumininu Oguntola, Xiaqing Pan, Penny Peng, Shraman Pramanick, Merey Ramazanova, Fiona Ryan, Wei Shan,
  Kiran Somasundaram, Chenan Song, Audrey Southerland, Masatoshi Tateno, Huiyu Wang, Yuchen Wang, Takuma Yagi, Mingfei Yan, Xitong Yang, Zecheng Yu, Shengxin~Cindy Zha, Chen Zhao, Ziwei Zhao, Zhifan Zhu, Jeff Zhuo, Pablo Arbelaez, Gedas Bertasius, David Crandall, Dima Damen, Jakob Engel, Giovanni~Maria Farinella, Antonino Furnari, Bernard Ghanem, Judy Hoffman, C.~V. Jawahar, Richard Newcombe, Hyun~Soo Park, James~M. Rehg, Yoichi Sato, Manolis Savva, Jianbo Shi, Mike~Zheng Shou, and Michael Wray.
\newblock Ego-exo4d: Understanding skilled human activity from first- and third-person perspectives, 2024.

\bibitem[Hassan et~al.(2019)Hassan, Choutas, Tzionas, and Black]{PROX:2019}
Mohamed Hassan, Vasileios Choutas, Dimitrios Tzionas, and Michael~J. Black.
\newblock Resolving {3D} human pose ambiguities with {3D} scene constraints.
\newblock In \emph{International Conference on Computer Vision}, pages 2282--2292, 2019.

\bibitem[Hassan et~al.(2021)Hassan, Ceylan, Villegas, Saito, Yang, Zhou, and Black]{SAMP2021}
Mohamed Hassan, Duygu Ceylan, Ruben Villegas, Jun Saito, Jimei Yang, Yi Zhou, and Michael~J Black.
\newblock Stochastic scene-aware motion prediction.
\newblock In \emph{Proceedings of the IEEE/CVF International Conference on Computer Vision}, pages 11374--11384, 2021.

\bibitem[Jiang et~al.({\natexlab{a}})Jiang, He, Wang, Li, Chen, Huang, and Zhu]{LINGOMOTION2024}
Nan Jiang, Zimo He, Zi Wang, Hongjie Li, Yixin Chen, Siyuan Huang, and Yixin Zhu.
\newblock Autonomous {{Character-Scene Interaction Synthesis}} from {{Text Instruction}}, {\natexlab{a}}.

\bibitem[Jiang et~al.({\natexlab{b}})Jiang, Zhang, Li, Ma, Wang, Chen, Liu, Zhu, and Huang]{TRUMANS2024}
Nan Jiang, Zhiyuan Zhang, Hongjie Li, Xiaoxuan Ma, Zan Wang, Yixin Chen, Tengyu Liu, Yixin Zhu, and Siyuan Huang.
\newblock Scaling {{Up Dynamic Human-Scene Interaction Modeling}}, {\natexlab{b}}.

\bibitem[Jiang et~al.(2024)Jiang, He, Wang, Li, Chen, Huang, and Zhu]{jiang2024autonomous}
Nan Jiang, Zimo He, Zi Wang, Hongjie Li, Yixin Chen, Siyuan Huang, and Yixin Zhu.
\newblock Autonomous character-scene interaction synthesis from text instruction.
\newblock In \emph{SIGGRAPH Asia 2024 Conference Papers}, pages 1--11, 2024.

\bibitem[Li et~al.(2024)Li, Zhao, Zhang, Lyu, Dusmanu, Zhang, Pollefeys, and Tang]{li2024egogen}
Gen Li, Kaifeng Zhao, Siwei Zhang, Xiaozhong Lyu, Mihai Dusmanu, Yan Zhang, Marc Pollefeys, and Siyu Tang.
\newblock {EgoGen: An Egocentric Synthetic Data Generator}.
\newblock In \emph{IEEE Conference on Computer Vision and Pattern Recognition (CVPR)}, 2024.

\bibitem[Mahmood et~al.(2019)Mahmood, Ghorbani, Troje, Pons-Moll, and Black]{Mahmood2019-bi}
Naureen Mahmood, Nima Ghorbani, Nikolaus~F Troje, Gerard Pons-Moll, and Michael~J Black.
\newblock {AMASS: A}rchive of motion capture as surface shapes.
\newblock In \emph{{International Conference on Computer Vision ({ICCV})}}, 2019.

\bibitem[Mnih et~al.(2013)Mnih, Kavukcuoglu, Silver, Graves, Antonoglou, Wierstra, and Riedmiller]{qlearning2013}
Volodymyr Mnih, Koray Kavukcuoglu, David Silver, Alex Graves, Ioannis Antonoglou, Daan Wierstra, and Martin Riedmiller.
\newblock Playing atari with deep reinforcement learning.
\newblock \emph{arXiv preprint arXiv:1312.5602}, 2013.

\bibitem[Pavlakos et~al.(2019)Pavlakos, Choutas, Ghorbani, Bolkart, Osman, Tzionas, and Black]{smplifyPP}
Georgios Pavlakos, Vasileios Choutas, Nima Ghorbani, Timo Bolkart, Ahmed A.~A. Osman, Dimitrios Tzionas, and Michael~J. Black.
\newblock Expressive body capture: {3D} hands, face, and body from a single image.
\newblock In \emph{{Computer Vision and Pattern Recognition (CVPR)}}, pages 10975--10985, 2019.

\bibitem[Peng et~al.(2021)Peng, Ma, Abbeel, Levine, and Kanazawa]{Peng2021-vr}
Xue~Bin Peng, Ze Ma, Pieter Abbeel, Sergey Levine, and Angjoo Kanazawa.
\newblock Amp: adversarial motion priors for stylized physics-based character control.
\newblock \emph{{Transactions on Graphics (TOG)}}, 2021.

\bibitem[Petrovich et~al.(2022)Petrovich, Black, and Varol]{Petrovich2022-ie}
Mathis Petrovich, Michael~J Black, and Gül Varol.
\newblock Temos: Generating diverse human motions from textual descriptions.
\newblock In \emph{{European Conference on Computer Vision (ECCV)}}, 2022.

\bibitem[Punnakkal et~al.(2021)Punnakkal, Chandrasekaran, Athanasiou, Quiros-Ramirez, and Black]{BABEL:CVPR:2021}
Abhinanda~R. Punnakkal, Arjun Chandrasekaran, Nikos Athanasiou, Alejandra Quiros-Ramirez, and Michael~J. Black.
\newblock {BABEL}: Bodies, action and behavior with english labels.
\newblock In \emph{Proceedings IEEE/CVF Conf.~on Computer Vision and Pattern Recognition (CVPR)}, pages 722--731, 2021.

\bibitem[Schaul et~al.(2015)Schaul, Quan, Antonoglou, and Silver]{schaul2015prioritized}
Tom Schaul, John Quan, Ioannis Antonoglou, and David Silver.
\newblock Prioritized experience replay.
\newblock \emph{arXiv preprint arXiv:1511.05952}, 2015.

\bibitem[Schulman et~al.(2017)Schulman, Wolski, Dhariwal, Radford, and Klimov]{schulman2017proximal}
John Schulman, Filip Wolski, Prafulla Dhariwal, Alec Radford, and Oleg Klimov.
\newblock Proximal policy optimization algorithms.
\newblock \emph{arXiv preprint arXiv:1707.06347}, 2017.

\bibitem[Starke et~al.(2019)Starke, Zhang, Komura, and Saito]{Starke2019-bl}
Sebastian Starke, He Zhang, Taku Komura, and Jun Saito.
\newblock Neural state machine for character-scene interactions.
\newblock \emph{{Transactions on Graphics (TOG)}}, 2019.

\bibitem[Straub et~al.(2019)Straub, Whelan, Ma, Chen, Wijmans, Green, Engel, Mur-Artal, Ren, Verma, Clarkson, Yan, Budge, Yan, Pan, Yon, Zou, Leon, Carter, Briales, Gillingham, Mueggler, Pesqueira, Savva, Batra, Strasdat, Nardi, Goesele, Lovegrove, and Newcombe]{straub2019replica}
Julian Straub, Thomas Whelan, Lingni Ma, Yufan Chen, Erik Wijmans, Simon Green, Jakob~J. Engel, Raul Mur-Artal, Carl Ren, Shobhit Verma, Anton Clarkson, Mingfei Yan, Brian Budge, Yajie Yan, Xiaqing Pan, June Yon, Yuyang Zou, Kimberly Leon, Nigel Carter, Jesus Briales, Tyler Gillingham, Elias Mueggler, Luis Pesqueira, Manolis Savva, Dhruv Batra, Hauke~M. Strasdat, Renzo~De Nardi, Michael Goesele, Steven Lovegrove, and Richard Newcombe.
\newblock The replica dataset: A digital replica of indoor spaces, 2019.

\bibitem[Taheri et~al.(2020)Taheri, Ghorbani, Black, and Tzionas]{GRAB:2020}
Omid Taheri, Nima Ghorbani, Michael~J. Black, and Dimitrios Tzionas.
\newblock {GRAB}: {A} dataset of whole-body human grasping of objects.
\newblock In \emph{European Conference on Computer Vision ({ECCV})}, 2020.

\bibitem[Tessler et~al.(2024)Tessler, Guo, Nabati, Chechik, and Peng]{tessler2024maskedmimic}
Chen Tessler, Yunrong Guo, Ofir Nabati, Gal Chechik, and Xue~Bin Peng.
\newblock Maskedmimic: Unified physics-based character control through masked motion inpainting.
\newblock \emph{arXiv preprint arXiv:2409.14393}, 2024.

\bibitem[Tevet et~al.(2022)Tevet, Raab, Gordon, and Shafir]{Tevet2022-ab}
Guy Tevet, Sigal Raab, Brian Gordon, and Shafir.
\newblock Human motion diffusion model.
\newblock \emph{{arXiv}:2209.14916}, 2022.

\bibitem[Tevet et~al.(2024)Tevet, Raab, Cohan, Reda, Luo, Peng, Bermano, and van~de Panne]{tevet2024closd}
Guy Tevet, Sigal Raab, Setareh Cohan, Daniele Reda, Zhengyi Luo, Xue~Bin Peng, Amit~H Bermano, and Michiel van~de Panne.
\newblock Closd: Closing the loop between simulation and diffusion for multi-task character control.
\newblock \emph{arXiv preprint arXiv:2410.03441}, 2024.

\bibitem[Ugrinovic et~al.(2024)Ugrinovic, Lucas, Baradel, Weinzaepfel, Rogez, and Moreno-Noguer]{PURPOSER2024}
Nicolas Ugrinovic, Thomas Lucas, Fabien Baradel, Philippe Weinzaepfel, Gr{\'e}gory Rogez, and Francesc Moreno-Noguer.
\newblock Purposer: Putting human motion generation in context.
\newblock In \emph{2024 International Conference on 3D Vision (3DV)}, pages 1310--1319. IEEE, 2024.

\bibitem[Van~Hasselt et~al.(2016)Van~Hasselt, Guez, and Silver]{hado2016double}
Hado Van~Hasselt, Arthur Guez, and David Silver.
\newblock Deep reinforcement learning with double q-learning.
\newblock In \emph{Proceedings of the AAAI conference on artificial intelligence}, 2016.

\bibitem[Wang et~al.(2020)Wang, Arti{\`e}res, Chen, and Denoyer]{wang2020adversarial}
Qi Wang, Thierry Arti{\`e}res, Mickael Chen, and Ludovic Denoyer.
\newblock Adversarial learning for modeling human motion.
\newblock \emph{The Visual Computer}, 36\penalty0 (1):\penalty0 141--160, 2020.

\bibitem[Wang et~al.(2022)Wang, Chen, Liu, Zhu, Liang, and Huang]{wang2022humanise}
Zan Wang, Yixin Chen, Tengyu Liu, Yixin Zhu, Wei Liang, and Siyuan Huang.
\newblock Humanise: Language-conditioned human motion generation in 3d scenes.
\newblock \emph{Advances in Neural Information Processing Systems}, 35:\penalty0 14959--14971, 2022.

\bibitem[Yang et~al.(2025)Yang, Yang, Zhou, Chen, Zhang, Du, and Gan]{yang20253dmem3dscenememory}
Yuncong Yang, Han Yang, Jiachen Zhou, Peihao Chen, Hongxin Zhang, Yilun Du, and Chuang Gan.
\newblock 3d-mem: 3d scene memory for embodied exploration and reasoning, 2025.

\bibitem[Yao et~al.(2024)Yao, Song, Zhou, Ao, Chen, and Liu]{yao2024moconvq}
Heyuan Yao, Zhenhua Song, Yuyang Zhou, Tenglong Ao, Baoquan Chen, and Libin Liu.
\newblock Moconvq: Unified physics-based motion control via scalable discrete representations.
\newblock \emph{ACM Transactions on Graphics (TOG)}, 43\penalty0 (4):\penalty0 1--21, 2024.

\bibitem[Yi et~al.(2023)Yi, Huang, Tripathi, Hering, Thies, and Black]{yi2023mime}
Hongwei Yi, Chun-Hao~P. Huang, Shashank Tripathi, Lea Hering, Justus Thies, and Michael~J. Black.
\newblock {MIME}: Human-aware {3D} scene generation.
\newblock In \emph{IEEE/CVF Conf.~on Computer Vision and Pattern Recognition (CVPR)}, pages 12965--12976, 2023.

\bibitem[Zhang et~al.(2024)Zhang, Starke, Guzov, Zhang, P{\'e}rez-Pellitero, and Pons-Moll]{zhang2024scenic}
Xiaohan Zhang, Sebastian Starke, Vladimir Guzov, Zhensong Zhang, Eduardo P{\'e}rez-Pellitero, and Gerard Pons-Moll.
\newblock Scenic: Scene-aware semantic navigation with instruction-guided control.
\newblock \emph{arXiv preprint arXiv:2412.15664}, 2024.

\bibitem[Zhang and Tang(2022)]{Zhang_2022_CVPR}
Yan Zhang and Siyu Tang.
\newblock The wanderings of odysseus in 3{D} scenes.
\newblock In \emph{{Computer Vision and Pattern Recognition (CVPR)}}, 2022.

\bibitem[Zhao et~al.()Zhao, Li, and Tang]{DART2024}
Kaifeng Zhao, Gen Li, and Siyu Tang.
\newblock {{DART}}: {{A Diffusion-Based Autoregressive Motion Model}} for {{Real-Time Text-Driven Motion Control}}.

\bibitem[Zhao et~al.(2023)Zhao, Zhang, Wang, Beeler, , and Tang]{Zhao2023Synthesizing}
Kaifeng Zhao, Yan Zhang, Shaofei Wang, Thabo Beeler, , and Siyu Tang.
\newblock Synthesizing diverse human motions in 3d indoor scenes.
\newblock In \emph{International conference on computer vision (ICCV)}, 2023.

\end{thebibliography}
}


\clearpage
\appendix
{\noindent\LARGE\textbf{Supplementary Material}}
\newline
\renewcommand{\thefigure}{S.\arabic{figure}}
\renewcommand{\thetable}{S.\arabic{table}}
\renewcommand{\theequation}{S.\arabic{equation}}
\setcounter{figure}{0}
\setcounter{table}{0}
\setcounter{equation}{0}

\section{Q-Learning Training Parameters}
\label{sec:sup:qlearning}

In this section, we provide the details of the Q-learning training parameters used in our experiments. The key hyperparameters are as follows:

\begin{table}[h]
    \centering
    \begin{tabular}{lc}
        \toprule
        Parameter & Value \\
        \midrule
        Learning Rate & $1e^{-3}$ \\
        Discount Factor & $0.99$ \\
        Exploration Rate (Initial) & $1$ \\
        Exploration Rate (Final) & $0.1@2K$ steps \\
        Batch Size & $64$ \\
        Replay Buffer Size & $20000$ \\
        Target Update Frequency & $500$ \\
        Double Q-learning & $True$ \\
        Training steps & $30K$ \\
        \bottomrule
    \end{tabular}
    \caption{Q-learning hyperparameters.}
    \label{tab:hyperparameters}
\end{table}

\section{Visualization of the Scenes}

\begin{figure*}[!h]
\centerline{\includegraphics[width=\textwidth]{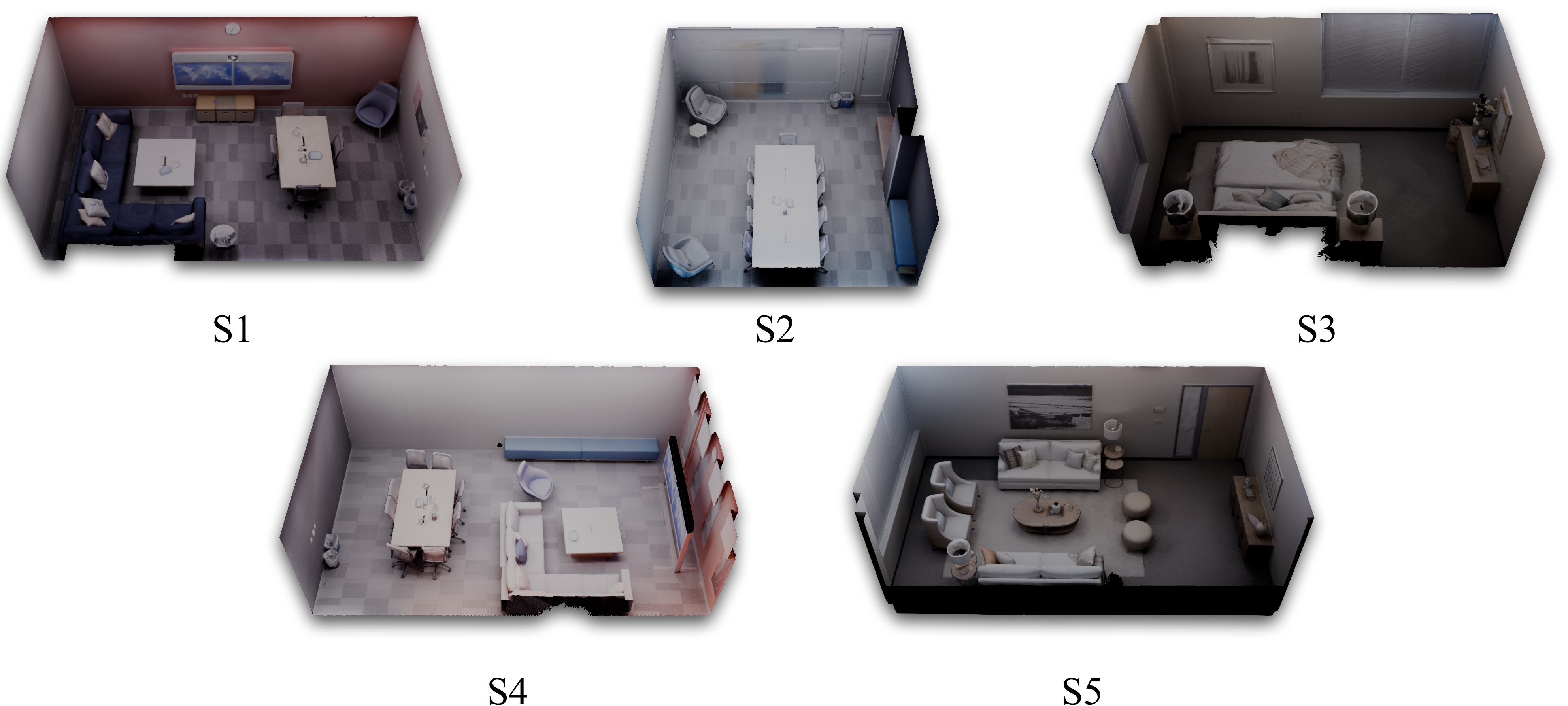}}
\vspace{-0.1in}
    \caption{The five scenes we use in training and testing of \methodname.}
    \label{sup:fig:scenes}
    
\end{figure*}

We present visualizations of the five different scenes used in our experiments (see Fig.~\ref{sup:fig:scenes}). These environments provide diverse challenges for the learning agent. From more open areas that require long-term planning like $S2$, to cluttered rooms like $S1, S4, S5$ where precise navigation is needed 

\section{More Qualitative Results}

\begin{figure*}[!h]
\centerline{\includegraphics[width=\textwidth]{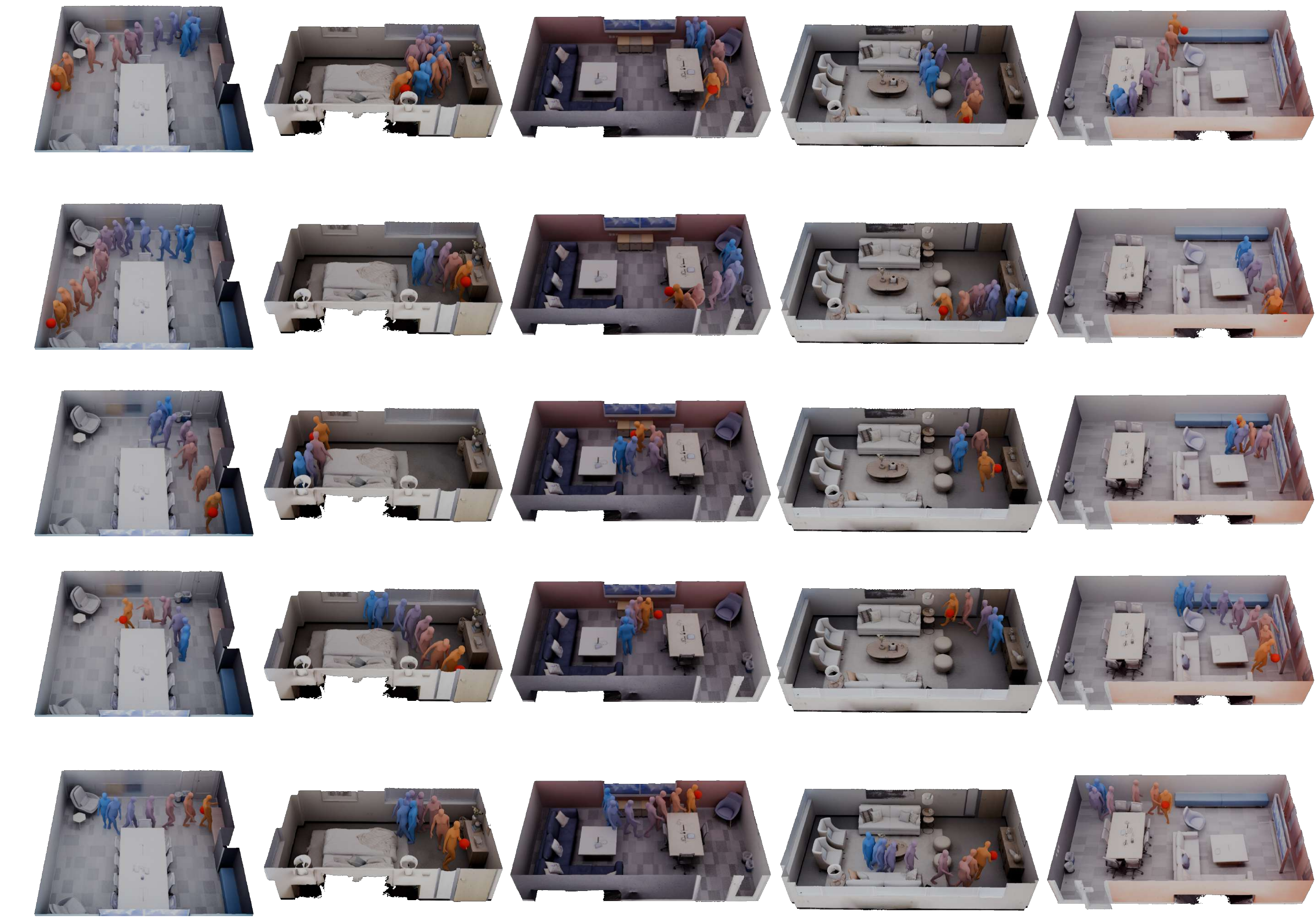}}
\vspace{-0.1in}
    \caption{More qualitative results of \methodname.}
    \label{sup:fig:morequals}
\end{figure*}

In Fig.~\ref{sup:fig:morequals} we present additional qualitative results to further demonstrate the effectiveness of our approach.
\clearpage



\end{document}